\documentclass[letterpaper]{article} 
\usepackage{aaai23}  
\usepackage{times}  
\usepackage{helvet}  
\usepackage{courier}  
\usepackage[hyphens]{url}  
\usepackage{graphicx} 
\usepackage{natbib}  
\usepackage{caption} 
\usepackage{algorithmicx}
\usepackage{algorithm}
\usepackage{algpseudocode}
\usepackage{booktabs}
\usepackage{amsmath,amsthm,amsfonts,amssymb}
\usepackage{bm}
\usepackage{multirow}
\usepackage{subfigure}
\usepackage{newfloat}
\usepackage{listings}
\DeclareCaptionStyle{ruled}{labelfont=normalfont,labelsep=colon,strut=off} 
\floatstyle{ruled}
\newfloat{listing}{tb}{lst}{}
\floatname{listing}{Listing}

\newtheorem{thm}{Theorem}

\newcommand{\argmin}{\mathop{\arg\min}}

\title{Partial-Label Regression}
\author{
Xin Cheng\textsuperscript{\rm 1}, Deng-Bao Wang\textsuperscript{\rm 2}, Lei Feng\textsuperscript{\rm 1}\thanks{Corresponding author: Lei Feng $<$lfeng@cqu.edu.cn$>$.}, Min-Ling Zhang\textsuperscript{\rm 2}, Bo An\textsuperscript{\rm 3}
}
\affiliations{
    \textsuperscript{\rm 1}College of Computer Science, Chongqing University, Chongqing, China\\
    \textsuperscript{\rm 2}School of Computer Science and Engineering, Southeast University, Nanjing, China\\
    \textsuperscript{\rm 3}School of Computer Science and Engineering, Nanyang Technological University, Singapore\\
    xincheng@stu.cqu.edu.cn, lfeng@cqu.edu.cn, \{wangdb, zhangml\}@seu.edu.cn, boan@ntu.edu.sg
}

\usepackage{bibentry}

\begin{document}

\maketitle
\begin{abstract}
Partial-label learning is a popular weakly supervised learning setting that allows each training example to be annotated with a set of candidate labels. Previous studies on partial-label learning only focused on the classification setting where candidate labels are all discrete, which cannot handle continuous labels with real values. In this paper, we provide the first attempt to investigate partial-label regression, where each training example is annotated with a set of real-valued candidate labels. To solve this problem, we first propose a simple baseline method that takes the average loss incurred by candidate labels as the predictive loss. The drawback of this method lies in that the loss incurred by the true label may be overwhelmed by other false labels. To overcome this drawback, we propose an identification method that takes the least loss incurred by candidate labels as the predictive loss. We further improve it by proposing a progressive identification method to differentiate candidate labels using progressively updated weights for incurred losses. We prove that the latter two methods are model-consistent and provide convergence analyses. Our proposed methods are theoretically grounded and can be compatible with any models, optimizers, and losses. Experiments validate the effectiveness of our proposed methods. 
\end{abstract}

\section{Introduction}
Due to the difficulty of collecting strong supervision information (i.e., fully labeled datasets) in some real-world scenarios, many weakly supervised learning settings were investigated to deal with weak supervision information. Typical weakly supervised learning settings include semi-supervised learning~\cite{Chapelle2006semi,sohn2020fixmatch}, noisy-label learning \cite{liu2015classification,malach2017decoupling,patrini2017making}, and positive-unlabeled learning \cite{elkan2008learning,niu2016theoretical,Kiryo2017Positive}, and multiple-instance learning \cite{maron1997framework,andrews2002support}.

In recent years, another weakly supervised learning setting called \emph{partial-label learning} (PLL) \cite{cour2011learning} has received much attention from the machine learning and data mining communities. In PLL, each training example is annotated with a set of candidate labels, only one of which is the true label. Due to the massive label ambiguity and noise in data annotation tasks, PLL has been increasingly used in many real-world applications, such as web mining~\cite{luo2010learning}, multimedia content analysis~\cite{zeng2013learning}, and automatic image annotations~\cite{chen2018learning}.

The major challenge of PLL lies in label ambiguity, as the true label is concealed in the candidate label set and not directly accessible to the learning algorithm. To tackle this problem, many PLL methods have been proposed. These methods achieved satisfactory performance by using appropriate techniques, such as the expectation-maximization algorithm \cite{jin2003learning,wang2022pico}, the maximum margin criterion \cite{nguyen2008classification}, metric learning \cite{gong2021discriminative,liu2018metric}, the manifold regularization \cite{zhang2015solving,zhang2016partial,gong2018regularization,wang2019adaptive}, and the self-training strategy \cite{feng2019partial,lv2020progressive,feng2020provably,wen2021leveraged}.

Despite the effectiveness of previous PLL methods, they only focused on the classification setting where candidate labels are all discrete, which cannot handle continuous labels with real values. In many real-world scenarios, regression tasks that learn with real-valued labels can be commonly encountered.
However, \emph{how to learn an effective regression model with a set of real-valued candidate labels is an open problem that still remains unexplored}.

In this paper, we provide the first attempt to investigate \emph{partial-label regression} (PLR), where each training example is annotated with a set of real-valued candidate labels. In order to solve the PLR problem, we make the following contributions:
\begin{itemize}
\item We propose a simple baseline method that takes the average loss incurred by candidate labels as the predictive loss to be minimized for model training.
\item We propose an identification method that takes the least loss incurred by candidate labels as the predictive loss to be minimized for model training. 
\item We propose a progressive identification method that differentiates candidate labels by associating their incurred losses with progressively updated weights.
\item We theoretically show that the identification method and the progressive identification method are \emph{model-consistent}, which indicates that the learned model converges to the optimal model.
\end{itemize}
\section{Preliminaries}
In this section, we briefly introduce preliminary knowledge and studies that are related to our PLR problem.

\paragraph{Partial-Label Learning.} PLL is a popular weakly supervised classification problem \cite{gong2021top,gong2022partial}, where each training example is annotated with a set of discrete candidate labels, only one of which is the true label. Given such training data, PLL aims to construct a multi-class classifier that predicts the label of unseen test data as accurately as possible. The key challenge of PLL is that the false labels in the candidate label set would mislead the model training process. To tackle this problem, many efforts have been made to disambiguate the ambiguous candidate label set. For example, some methods aim at identifying the true label from the candidate label set by using appropriate techniques such as the maximum margin criterion \cite{nguyen2008classification,yu2016maximum} or class activation value \cite{zhang2022exploiting}. Some iterative methods \cite{feng2019partial,lv2020progressive,feng2020provably} characterize the different contributions of different candidate labels by using confidence scores and iteratively update the confidence score of each candidate label. Although these PLL methods have achieved satisfactory performance, they only focused on the classification setting where candidate labels are all discrete, which cannot handle continuous labels with real values. To solve this problem, our work focuses on learning a regression model with real-valued candidate labels, which would be more challenging than PLL because the label space becomes infinite when candidate labels are real-valued. 

\paragraph{Regression.} For the ordinary regression problem, let the feature space be $\mathcal{X}\in\mathbb{R}^d$ and the label space be $\mathcal{Y}\in\mathbb{R}$. Let us denote by $(\bm{x},y)$ an example including an instance $x$ and a real-valued label $y$. Each example $(\bm{x},y)\in\mathcal{X}\times\mathcal{Y}$ is assumed to be independently sampled from an unknown data distribution with probability density $p(\bm{x},y)$. For the regression task, we aim to learn a regression model $f:\mathcal{X}\mapsto\mathbb{R}$ that minimizes the following expected risk:
\begin{gather}
\label{expected_regression}
R(f)=\mathbb{E}_{p(\bm{x},y)}[\ell(f(\bm{x}),y)],
\end{gather}
where $\mathbb{E}_{p(\bm{x},y)}$ denotes the expectation over the data distribution $p(\bm{x},y)$ and $\ell:\mathbb{R}\times\mathbb{R}\mapsto\mathbb{R}_+$ is a conventional loss function (such as mean squared error and mean absolute error) for regression, which measures how well a model estimates a given real-valued label. As $p(\boldsymbol{x},y)$ is not available and we are given only a number of training examples $\{\boldsymbol{x}_i,y_i\}_{i=1}^n$ that are independently drawn from $p(\boldsymbol{x},y)$, a common strategy is to minimize the empirical risk
\begin{gather}
\label{empirical_regression}
\widehat{R}(f) = \frac{1}{n}\sum\nolimits_{i=1}^n\ell(f(\boldsymbol{x}_i),y_i),
\end{gather} 
which is called \emph{empirical risk minimization} \cite{vapnik1999overview}. It can be clearly seen that such a supervised regression method can only deal with fully labeled data where true labels are provided. In our work, we provide the first attempt to investigate \emph{a novel weakly supervised regression problem called partial-label regression}, where each training example is annotated with a set of real-valued candidate labels.


\section{The Proposed Methods}
In this section, we present effective methods to train a regression model from data with a set of real-valued candidate labels. We first propose a simple baseline method that averages the contributions of all the candidate labels. However, this intuitive method does not differentiate the true label for model training, and thus may cause the training process to be misled by false labels in the candidate label set. To overcome this drawback, we further propose two theoretically grounded methods, where one directly identifies the true label with the least loss for model training and the other progressively identifies the true label by associating the loss of each candidate label with properly updated weights.
\paragraph{Notations.} Suppose the training set for PLR is denoted by $\{(\bm{x}_i,S_i)\}_{i=1}^n$, where $S_i$ represents the set of real-valued candidate labels assigned to the instance $\bm{x}_i\in\mathcal{X}$, and each training example $(\bm{x}_i, S_i)$ is assumed to be sampled from an unknown data distribution with probability density $p(\bm{x}, S)$. In the PLR setting, the true label $y_i\in\mathcal{Y}$ of the instance $\bm{x}_i$ is always contained in its candidate label set $S_i$, i.e., $y_i\in S_i$. In addition, we also assume that the PLR setting satisfies the condition that the \emph{ambiguity degree} \cite{cour2011learning} is less than 1, which is defined as
\begin{gather}
\nonumber
\gamma = \sup_{(\bm{x},y)\sim p(\bm{x},y),(\bm{x},S)\sim p(\bm{x},S),y' \in \mathcal{Y},y' \neq y} p(y' \in S).
\end{gather}
As shown in the above equation, the ambiguity degree $\gamma$ is the maximum probability of an incorrect label $y^\prime$ being contained in the candidate label set $S$ (co-occurring with the true label $y$). If $\gamma=1$, we cannot differentiate the true label $y$ from the false label $y^\prime$, since they always appear in the same candidate label set.

\subsection{The Average Method}
An intuitive method to solve the PLR problem is to treat each candidate label equally and average the incurred loss of each candidate label:
\begin{gather}
\label{avg}
\ell_{\mathrm{avg}}(f(\bm{x}), S) = \frac{1}{|S|}\sum\nolimits_{y\in S}\ell(f(\bm{x}), y).
\end{gather}
In the above equation, $\ell_{\mathrm{avg}}(\bm{x}, S)$ can be taken as a loss function specially designed for the PLR problem. In Eq.~(\ref{avg}), we take into account the influence of each candidate label and regard the averaged loss as the predictive loss on the PLR example $(\bm{x},S)$. The average method is intuitive, while the major drawback of this method lies in that the loss incurred by the true label may be overwhelmed by other false labels in the candidate label set.

\subsection{The Identification Method}
We can find that the drawback of the average method comes from that it does not differentiate the true label from the set of real-valued candidate labels. Apart from the true label, there are normally multiple false labels in the candidate label set, hence these false labels may dominate the model training process and thus have huge negative impacts on the learned model. To overcome this drawback, we propose an identification method, which regards the candidate label with the least loss as the true label and only considers the least loss of the identified pseudo label as the predictive loss:
\begin{gather}
\label{idt}
\ell_{\mathrm{min}}(f(\bm{x}), S)=\min\nolimits_{y\in S}\ell(f(\bm{x}), y).
\end{gather}
Then, the expected risk and the empirical risk with our proposed identification method (i.e., $\ell_{\mathrm{min}}$) can be represented as follows:
\begin{align}
\nonumber
R_{\mathrm{min}}(f)&=\mathbb{E}_{p(\bm{x}, S)}[\ell_{\mathrm{min}}(f(\bm{x}),S)],\\
\label{min_empirical_risk}
\widehat{R}_{\mathrm{min}}(f)&=\frac{1}{n}\sum\nolimits_{i=1}^n[\ell_{\mathrm{min}}(f(\bm{x}_i),S_i)].
\end{align}
By directly minimizing the derived empirical risk $\widehat{R}_{\mathrm{min}}(f)$, we can learn an effective regression model from training data with only real-valued candidate labels. The key idea of the identification method lies in that the model has its own ability to identify the true label through the training process. This idea is quite similar to the small-loss selection strategy used in the noisy-label learning problem \cite{han2018co,wei2020combating}.
\paragraph{Model Consistency.} We demonstrate that the identification method is \emph{model-consistent}. That is, the model learned by the identification method from data with real-valued candidate labels converges to the optimal model learned from fully supervised data. It is noteworthy that the hypothesis space $\mathcal{F}$ is commonly assumed to be powerful enough \cite{lv2020progressive}, hence the optimal model in the hypothesis space (i.e., $f^\star = \argmin_{f\in\mathcal{F}}{R}(f)$) makes the optimal risk equal to 0 (i.e., $R(f^\star)=0$). We also adopt this assumption throughout this paper.
\begin{thm}
\label{thm1}
The model $f^\star_{\mathrm{min}}=\argmin_{f\in\mathcal{F}}R_{\mathrm{min}}(f)$ learned by the identification method is equivalent to the optimal model $f^\star = \argmin_{f\in\mathcal{F}}{R}(f)$.
\end{thm}
Theorem \ref{thm1} shows that the optimal model learned from fully labeled data can be identified by our identification method given only data with real-valued candidate labels.
\paragraph{Convergence Analysis.} Here, we provide a convergence analysis for the above identification method, which shows that the model $\widehat{f}_{\mathrm{min}}=\argmin_{f\in\mathcal{F}}\widehat{R}_{\mathrm{min}}(f)$ (empirically learned from only data with real-valued candidate labels by using our identification method) can converge to the optimal model $f^\star$. Given such a convergence analysis, we can observe that the identification method could benefit from the increasing number of training data with real-valued candidate labels. To ensure that $\widehat{f}_{\mathrm{min}}$ converges to $f^\star$, we can show that $R_{\mathrm{min}}(\widehat{f}_{\mathrm{min}})$ converges to $R_{\mathrm{min}}(f^\star)$. Since we have proved the model consistency of the identification method (i.e., $f^\star=f^\star_{\mathrm{min}}$), we can turn to show that $R_{\mathrm{min}}(\widehat{f}_{\mathrm{min}})$ converges to $R_{\mathrm{min}}(f^\star_{\mathrm{min}})$.
\begin{thm}
\label{thm2}
Suppose the pseudo-dimensions of $\{\bm{x}\mapsto \ell(f(\bm{x}),y)\mid f\in\mathcal{F}, y\in\mathcal{Y}\}$ is finite and there exists a constant $M\leq\infty$ such that $|\ell(f(\bm{x}),y)|\leq M$ for all $(\bm{x},y)\in\mathcal{X}\times\mathcal{Y}$ and $f\in\mathcal{F}$. Then, the estimation error $R_{\mathrm{min}}(\widehat{f}_{\mathrm{min}})-R_{\mathrm{min}}(f^\star)$ would decrease to zero in the order $\mathcal{O}({1}/{\sqrt{n}})$, where $n$ is the number of data with real-valued candidate labels.
\end{thm}
Theorem \ref{thm2} demonstrates that the optimal model can be learned by the identification method when the number of training data for PLR approaches infinity. 

\subsection{The Progressive Identification Method}
We have introduced the average method and the identification method earlier in this section. The first one treats all the candidate targets equally and the second one focuses too much on a single candidate label, hence both of them fail consider different contributions of candidate labels. To remedy this issue, we further propose a progressive identification method that takes the weighted loss incurred by candidate labels as the predictive loss:
\begin{gather}
\nonumber
\ell_{\mathrm{wet}}(f(\bm{x}), S) = \sum\nolimits_{y\in S}w(\bm{x},y)\ell(f(\bm{x}),y),
\end{gather}
where $w(\bm{x},y)$ is a weighting function that describes the importance degree of the label $y$ to the instance $\bm{x}$. Hence in the PLR task, for each training instance $\bm{x}$, $w(\bm{x},y)$ is expected to satisfy the following conditions:
\begin{gather}
\label{candidate_constraint}
\forall y\in S, w(\bm{x},{y})\geq 0\ \mathrm{and}\
\sum\nolimits_{y\in S} w(\bm{x},y)=1.
\end{gather}
Eq.~(\ref{candidate_constraint}) implies that every candidate label may have an impact on model training, since each of them has the probability of being the true label while only one of them is the true label. It is noteworthy that by manually setting $w(\bm{x},y)$ to different values, the progressive identification method can recover the average method and the identification method. 

Then, the expected risk and the empirical risk with our proposed progressive identification method (i.e., $\ell_{\mathrm{wet}}$) can be represented as follows:
\begin{align}
\nonumber
R_{\mathrm{wet}}(f)&=\mathbb{E}_{p(\bm{x}, S)}[\ell_{\mathrm{wet}}(f(\bm{x}),S)],\\
\label{weighting_empirical_risk}
\widehat{R}_{\mathrm{wet}}(f)&=\frac{1}{n}\sum\nolimits_{i=1}^n[\ell_{\mathrm{wet}}(f(\bm{x}_i),S_i)],
\end{align}
By directly minimizing the derived empirical risk $\widehat{R}_{\mathrm{wet}}(f)$, we can learn an effective model from training data with only real-valued candidate labels.
\paragraph{Model Consistency.} We show that the progressive identification method is also model-consistent.
\begin{thm}
\label{thm3}
Suppose the optimal weighting function can be achieved: $\forall (\bm{x},y), w(\bm{x},y)=1$ and $w(\bm{x},y^\prime)= 0$ for $y^\prime\neq y$.
The model $f^\star_{\mathrm{wet}}=\argmin_{f\in\mathcal{F}}R_{\mathrm{wet}}(f)$ is equivalent to the optimal model $f^\star = \argmin_{f\in\mathcal{F}}{R}(f)$.
\end{thm}
Theorem \ref{thm3} shows that the optimal regression model learned from fully labeled data can be identified by the progressive identification method given only data with real-valued candidate labels.
\paragraph{Convergence Analysis.} Here, we also provide a convergence analysis for the identification method. We aim to show that the model $\widehat{f}_{\mathrm{wet}}=\argmin_{f\in\mathcal{F}}\widehat{R}_{\mathrm{wet}}(f)$ (empirically learned from only data with real-valued candidate labels by using our progressive identification method $\ell_{\mathrm{wet}}$) can converge to the optimal model $f^\star$.
\begin{thm}
\label{thm4}
With the same conditions in Theorem \ref{thm2}, the estimation error $R_{\mathrm{wet}}(\widehat{f}_{\mathrm{wet}})-R_{\mathrm{wet}}(f^\star)$ would decrease to zero in the order $\mathcal{O}({1}/{\sqrt{n}})$, where $n$ is the number of data with real-valued candidate labels.
\end{thm}
Theorem \ref{thm4} shows that the optimal model can also be learned by the progressive identification method when the number of training data for PLR approaches infinity. 

\paragraph{Weighting Function Design.} Here, we provide the discussion on the specific choice of the weighting function $w(\bm{x},y)$. Motivated by the key idea of the identification method, we also consider that the importance degrees of candidate labels can be reflected by the incurred losses. Specifically, if a candidate label has a smaller loss than other candidate labels, then a larger weight (importance degree) should be assigned to this candidate label. Besides, if the loss of a candidate label approaches 0, we consider this candidate label to be the true label, hence the weight assigned to this candidate label would approach 1, and the weights of other labels would approach 0. Based on this perspective, we design the weighting function as follows:
\begin{gather}
\label{weighting_design}
w(\bm{x},y) = \left\{\begin{matrix}
 \frac{\exp(\beta_2\cdot\ell(f(\bm{x}),y)^{-\beta_1})}{\sum_{y^\prime\in S}\exp(\beta_2\cdot\ell(f(\bm{x}),y^\prime)^{-\beta_1})}, &\mathrm{if}\ y\in S, \\
0, &\mathrm{otherwise}, \\
\end{matrix}\right.
\end{gather}
where $\beta_1$ and $\beta_2$ are two hyper-parameters. For the design of $w(\bm{x},y)$ in Eq.~(\ref{weighting_design}), we first compute a score of the candidate label $y$ by $\beta_2\cdot\ell(f(\bm{x}),y)^{-\beta_1}$, which implies that candidate labels with smaller losses would have larger weights. Then, we use the softmax function to normalize the scores of candidate labels and set the weights of non-candidate labels to zero. In this way, the design of $w(\bm{x},y)$ in Eq.~(\ref{weighting_design}) satisfies all the conditions in Eq.~(\ref{candidate_constraint}). It is noteworthy that such a design is not unique, and there may exist better designs. We leave the exploration of other designs of the weighting function $w(\bm{x},y)$ for future work.


\section{Experiments}
\subsection{Experimental Setup}
\paragraph{Datasets.}
We use seven widely used benchmark regression datasets including Abalone, Airfoil, Auto-mpg, Housing, Concrete, Power-plant, and Cpu-act. All of these datasets can be downloaded from the UCI Machine Learning Repository\footnote{https://archive.ics.uci.edu/}. For each dataset, we randomly split the original dataset into training, validation, and test sets by the proportions of 60\%, 20\%, and 20\%, respectively. For each instance (feature vector) $\bm{x}$, the min-max normalization is used for dimensions with a fixed range of values (i.e., $(\bm{x}-\min(\bm{x}))/(\max(\bm{x})-\min(\bm{x}))$, and the one-hot encoding is used to encode dimensions of the discrete type. The remaining continuous dimensions are standardized to have mean 0 and standard deviation 1. The characteristics of these datasets are reported in Table~\ref{tab1}. We repeat the sampling-and-training process 10 times on these datasets and record the mean squared error with standard deviation.
\begin{table}[!t]
\centering
\resizebox{0.45\textwidth}{!}{
\begin{tabular}{c|c|c|c|c}
\toprule
Dataset     & \# Train & \# Test & \# Validation & \# Features    \\ \hline
Abalone     & 2507     & 835     & 835           & 8 \\
Airfoil     & 903      & 300     & 300           & 5 \\
Auto-mpg    & 236      & 78      & 78            & 7  \\
Housing     & 304      & 101     & 101           & 13 \\
Concrete    & 618      & 206     & 206           & 8  \\
Power-plant & 5742     & 1913    & 1913          & 4  \\
Cpu-act     & 4916     & 1638    & 1638          & 21 \\
\bottomrule
\end{tabular}
}
\caption{Characteristics of the used benchmark datasets.}
\label{tab1}
\end{table}
\begin{table*}[!t]
\centering
\resizebox{1.00\textwidth}{!}{
\begin{tabular}{clcccccccc|cc}
\toprule
\multicolumn{2}{c}{\multirow{2}{*}{Datasets}}    & \multirow{2}{*}{$|\bar{S}|$} & \multirow{2}{*}{Supervised} & \multirow{2}{*}{AVGL-MSE} & \multirow{2}{*}{AVGL-MAE} & \multirow{2}{*}{AVGL-Huber} & \multirow{2}{*}{AVGV-MSE} & \multirow{2}{*}{AVGV-MAE} & \multirow{2}{*}{AVGV-Huber} & \multirow{2}{*}{IDent} & \multirow{2}{*}{PIDent} \\
\multicolumn{2}{c}{}                             &                       &                             &                           &                           &                             &                           &                           &                             &                     &                       \\ \hline
\multicolumn{2}{c}{\multirow{8}{*}{Abalone}}     & \multirow{2}{*}{2}    &                             & 9.72                      & 4.66                      & 4.68                        & 9.72                      & 7.36                      & 7.52                        & 4.62                & \textbf{4.55}         \\
\multicolumn{2}{c}{}                             &                       &                             & (0.77)                    & (0.30)                    & (0.29)                      & (0.78)                    & (0.92)                    & (1.05)                      & (0.33)              & \textbf{(0.21)}       \\
\multicolumn{2}{c}{}                             & \multirow{2}{*}{4}    &                             & 14.42                     & 5.04                      & 5.22                        & 14.08                     & 14.08                     & 14.04                       & 4.66                & \textbf{4.58}         \\
\multicolumn{2}{c}{}                             &                       & 4.66                        & (0.55)                    & (0.30)                    & (0.31)                      & (0.93)                    & (0.93)                    & (0.61)                      & (0.20)              & \textbf{(0.22)}       \\
\multicolumn{2}{c}{}                             & \multirow{2}{*}{8}    & (0.54)                      & 20.63                     & 7.76                      & 7.94                        & 22.68                     & 22.68                     & 21.28                       & \textbf{4.70}       & 4.71                  \\
\multicolumn{2}{c}{}                             &                       &                             & (2.13)                    & (0.82)                    & (0.85)                      & (3.88)                    & (3.88)                    & (2.94)                      & \textbf{(0.26)}     & (0.29)                \\
\multicolumn{2}{c}{}                             & \multirow{2}{*}{16}   &                             & 25.11                     & 13.77                     & 14.03                       & 26.71                     & 26.71                     & 26.61                       & \textbf{4.90}       & \textbf{4.90}         \\
\multicolumn{2}{c}{}                             &                       &                             & (0.65)                    & (0.71)                    & (0.77)                      & (1.37)                    & (1.37)                    & (2.63)                      & \textbf{(0.27)}     & \textbf{(0.35)}       \\ \hline
\multicolumn{2}{c}{\multirow{8}{*}{Airfoil}}     & \multirow{2}{*}{2}    &                             & 23.72                     & 16.66                     & 16.17                       & 23.73                     & 23.27                     & 22.99                       & 15.58               & \textbf{14.99}        \\
\multicolumn{2}{c}{}                             &                       &                             & (2.71)                    & (2.55)                    & (2.21)                      & (2.78)                    & (3.11)                    & (2.44)                      & (2.65)              & \textbf{(3.45)}       \\
\multicolumn{2}{c}{}                             & \multirow{2}{*}{4}    &                             & 30.79                     & 18.47                     & 18.56                       & 30.83                     & 30.67                     & 30.22                       & 16.23               & \textbf{16.10}        \\
\multicolumn{2}{c}{}                             &                       & 13.77                       & (3.99)                    & (3.08)                    & (3.04)                      & (3.94)                    & (2.68)                    & (2.92)                      & (3.71)              & \textbf{(2.97)}       \\
\multicolumn{2}{c}{}                             & \multirow{2}{*}{8}    & (3.27)                      & 37.19                     & 24.67                     & 24.40                       & 37.16                     & 39.24                     & 38.19                       & \textbf{17.81}      & 17.86                 \\
\multicolumn{2}{c}{}                             &                       &                             & (4.25)                    & (3.33)                    & (1.97)                      & (4.27)                    & (3.44)                    & (3.75)                      & \textbf{(3.49)}     & (3.13)                \\
\multicolumn{2}{c}{}                             & \multirow{2}{*}{16}   &                             & 42.90                     & 31.43                     & 30.81                       & 43.00                     & 45.38                     & 44.19                       & \textbf{23.41}      & 24.11                 \\
\multicolumn{2}{c}{}                             &                       &                             & (3.86)                    & (4.50)                    & (3.78)                      & (3.76)                    & (3.84)                    & (4.81)                      & \textbf{(5.08)}     & (2.98)                \\ \hline
\multicolumn{2}{c}{\multirow{8}{*}{Auto-mpg}}    & \multirow{2}{*}{2}    &                             & 21.69                     & 21.69                     & 10.03                       & 21.69                     & 17.73                     & 16.74                       & 9.07                & \textbf{8.64}         \\
\multicolumn{2}{c}{}                             &                       &                             & (3.50)                    & (3.50)                    & (2.57)                      & (3.50)                    & (4.35)                    & (4.93)                      & (1.74)              & \textbf{(1.84)}       \\
\multicolumn{2}{c}{}                             & \multirow{2}{*}{4}    &                             & 31.66                     & 31.66                     & 13.35                       & 31.67                     & 31.41                     & 30.40                       & 9.10                & \textbf{9.09}         \\
\multicolumn{2}{c}{}                             &                       & 8.77                        & (4.74)                    & (4.74)                    & (3.49)                      & (4.74)                    & (4.22)                    & (5.24)                      & (2.32)              & \textbf{(2.43)}       \\
\multicolumn{2}{c}{}                             & \multirow{2}{*}{8}    & (1.95)                      & 41.57                     & 17.85                     & 18.13                       & 41.57                     & 45.74                     & 43.65                       & \textbf{9.69}       & 10.10                 \\
\multicolumn{2}{c}{}                             &                       &                             & (3.62)                    & (1.97)                    & (2.24)                      & (3.62)                    & (6.16)                    & (4.93)                      & \textbf{(2.18)}     & (2.81)                \\
\multicolumn{2}{c}{}                             & \multirow{2}{*}{16}   &                             & 51.27                     & 32.11                     & 32.23                       & 51.27                     & 55.06                     & 54.42                       & 12.54               & \textbf{12.14}        \\
\multicolumn{2}{c}{}                             &                       &                             & (6.95)                    & (6.14)                    & (3.20)                      & (6.95)                    & (7.65)                    & (6.82)                      & (2.27)              & \textbf{(2.44)}       \\ \hline
\multicolumn{2}{c}{\multirow{8}{*}{Housing}}     & \multirow{2}{*}{2}    &                             & 33.82                     & 18.69                     & 17.37                       & 33.82                     & 32.79                     & 34.03                       & 16.18               & \textbf{15.55}        \\
\multicolumn{2}{c}{}                             &                       &                             & (6.96)                    & (4.64)                    & (4.63)                      & (6.93)                    & (8.45)                    & (8.14)                      & (3.30)              & \textbf{(3.89)}       \\
\multicolumn{2}{c}{}                             & \multirow{2}{*}{4}    &                             & 48.82                     & 26.78                     & 26.82                       & 48.80                     & 51.87                     & 53.67                       & 21.59               & \textbf{20.53}        \\
\multicolumn{2}{c}{}                             &                       & 14.48                       & (10.97)                   & (7.32)                    & (8.18)                      & (10.96)                   & (9.88)                    & (8.75)                      & (8.78)              & \textbf{(6.97)}       \\
\multicolumn{2}{c}{}                             & \multirow{2}{*}{8}    & (3.99)                      & 60.87                     & 38.18                     & 39.19                       & 60.90                     & 64.02                     & 62.23                       & 29.82               & \textbf{26.87}        \\
\multicolumn{2}{c}{}                             &                       &                             & (7.95)                    & (6.81)                    & (7.03)                      & (7.97)                    & (8.21)                    & (7.86)                      & (13.33)             & \textbf{(6.83)}       \\
\multicolumn{2}{c}{}                             & \multirow{2}{*}{16}   &                             & 75.30                     & 52.24                     & 53.06                       & 75.29                     & 84.18                     & 77.00                       & \textbf{39.88}      & 41.09                 \\
\multicolumn{2}{c}{}                             &                       &                             & (9.38)                    & (8.96)                    & (8.73)                      & (9.40)                    & (14.38)                   & (9.25)                      & \textbf{(19.69)}    & (15.24)               \\ \hline
\multicolumn{2}{c}{\multirow{8}{*}{Concrete}}    & \multirow{2}{*}{2}    &                             & 108.08                    & 46.66                     & 48.62                       & 108.23                    & 107.46                    & 106.00                      & 42.14               & \textbf{40.48}        \\
\multicolumn{2}{c}{}                             &                       &                             & (9.93)                    & (9.09)                    & (6.07)                      & (9.76)                    & (16.31)                   & (16.96)                     & (7.68)              & \textbf{(10.68)}      \\
\multicolumn{2}{c}{}                             & \multirow{2}{*}{4}    &                             & 151.05                    & 75.49                     & 80.38                       & 151.05                    & 172.01                    & 167.53                      & 45.61               & \textbf{44.87}        \\
\multicolumn{2}{c}{}                             &                       & 36.49                       & (15.19)                   & (22.58)                   & (17.05)                     & (15.17)                   & (20.88)                   & (21.83)                     & (5.10)              & \textbf{(7.30)}       \\
\multicolumn{2}{c}{}                             & \multirow{2}{*}{8}    & (3.08)                      & 195.40                    & 116.18                    & 118.65                      & 195.32                    & 216.01                    & 213.59                      & 72.09               & \textbf{63.79}        \\
\multicolumn{2}{c}{}                             &                       &                             & (14.12)                   & (25.99)                   & (10.85)                     & (14.06)                   & (26.26)                   & (18.63)                     & (22.53)             & \textbf{(14.13)}      \\
\multicolumn{2}{c}{}                             & \multirow{2}{*}{16}   &                             & 239.26                    & 186.94                    & 183.64                      & 239.64                    & 268.56                    & 259.05                      & 114.35              & \textbf{112.02}       \\
\multicolumn{2}{c}{}                             &                       &                             & (17.05)                   & (20.89)                   & (19.76)                     & (17.47)                   & (25.83)                   & (20.40)                     & (37.81)             & \textbf{(39.59)}      \\ \hline
\multicolumn{2}{c}{\multirow{8}{*}{Power-plant}} & \multirow{2}{*}{2}    &                             & 64.99                     & 23.67                     & 23.43                       & 64.99                     & 42.56                     & 42.68                       & 21.09               & \textbf{21.07}        \\
\multicolumn{2}{c}{}                             &                       &                             & (3.45)                    & (1.06)                    & (1.20)                      & (3.45)                    & (2.76)                    & (2.67)                      & (0.96)              & \textbf{(1.01)}       \\
\multicolumn{2}{c}{}                             & \multirow{2}{*}{4}    &                             & 104.61                    & 29.24                     & 29.47                       & 104.61                    & 98.36                     & 98.41                       & 21.29               & \textbf{21.28}        \\
\multicolumn{2}{c}{}                             &                       & 21.00                       & (6.27)                    & (1.43)                    & (1.17)                      & (6.27)                    & (5.64)                    & (5.79)                      & (1.13)              & \textbf{(1.19)}       \\
\multicolumn{2}{c}{}                             & \multirow{2}{*}{8}    & (1.06)                      & 153.07                    & 153.07                    & 48.49                       & 153.07                    & 167.96                    & 166.21                      & \textbf{21.21}      & 21.36                 \\
\multicolumn{2}{c}{}                             &                       &                             & (8.47)                    & (8.47)                    & (2.48)                      & (8.47)                    & (11.37)                   & (6.88)                      & \textbf{(1.06)}     & (0.89)                \\
\multicolumn{2}{c}{}                             & \multirow{2}{*}{16}   &                             & 204.32                    & 204.32                    & 105.45                      & 204.31                    & 225.97                    & 221.17                      & \textbf{21.31}      & 22.34                 \\
\multicolumn{2}{c}{}                             &                       &                             & (6.74)                    & (6.74)                    & (4.79)                      & (6.74)                    & (5.39)                    & (8.00)                      & \textbf{(0.89)}     & (1.99)                \\ \hline
\multicolumn{2}{c}{\multirow{8}{*}{Cpu-act}}     & \multirow{2}{*}{2}    &                             & 246.38                    & 9.84                      & 9.62                        & 245.68                    & 172.38                    & 169.26                      & \textbf{6.64}       & 6.91                  \\
\multicolumn{2}{c}{}                             &                       &                             & (15.60)                   & (1.04)                    & (0.79)                      & (15.12)                   & (18.19)                   & (18.16)                     & \textbf{(0.54)}     & (0.69)                \\
\multicolumn{2}{c}{}                             & \multirow{2}{*}{4}    &                             & 461.69                    & 33.15                     & 34.58                       & 461.05                    & 525.78                    & 520.64                      & \textbf{7.00}       & 7.38                  \\
\multicolumn{2}{c}{}                             &                       & 6.56                        & (13.66)                   & (5.75)                    & (7.32)                      & (12.63)                   & (17.09)                   & (14.98)                     & \textbf{(1.06)}     & (0.41)                \\
\multicolumn{2}{c}{}                             & \multirow{2}{*}{8}    & (0.86)                      & 730.38                    & 335.04                    & 333.92                      & 730.93                    & 858.55                    & 852.76                      & \textbf{7.48}       & 7.90                  \\
\multicolumn{2}{c}{}                             &                       &                             & (16.90)                   & (20.47)                   & (18.43)                     & (17.05)                   & (11.33)                   & (9.78)                      & \textbf{(1.32)}     & (0.64)                \\
\multicolumn{2}{c}{}                             & \multirow{2}{*}{16}   &                             & 972.73                    & 738.42                    & 735.89                      & 996.21                    & 1107.80                   & 1106.00                     & 9.48                & \textbf{9.11}         \\
\multicolumn{2}{c}{}                             &                       &                             & (29.53)                   & (26.71)                   & (24.00)                     & (46.98)                   & (54.23)                   & (25.97)                     & (2.77)              & \textbf{(1.93)}      
\\
\bottomrule
\end{tabular}
}
\caption{Test performance (mean squared error with standard deviation) of each method on the seven benchmark datasets training with the MLP model. The best performance is highlighted in bold.}
\label{tab2}
\end{table*}
\paragraph{Base Models.}
Since our proposed methods do not rely on a specific model, we train two types of base models including the linear model and three-layer multilayer perceptron (MLP) on the above benchmark datasets, to support the flexibility of our method on the choice of base models. The MLP model is a five-layer ($d$-20-30-10-1) neural network with the ReLU activation function. For both the linear model and the MLP model, we use the Adam optimization method \cite{kingma2015adam} with the batch size set to 256 and the number of training epochs set to 1000.

\begin{table*}[!t]
\centering
\resizebox{1.00\textwidth}{!}{
\begin{tabular}{clcccccccc|cc}
\toprule
\multicolumn{2}{c}{\multirow{2}{*}{Datasets}}    & \multirow{2}{*}{$|\bar{S}|$} & \multirow{2}{*}{Supervised} & \multirow{2}{*}{AVGL-MSE} & \multirow{2}{*}{AVGL-MAE} & \multirow{2}{*}{AVGL-Huber} & \multirow{2}{*}{AVGV-MSE} & \multirow{2}{*}{AVGV-MAE} & \multirow{2}{*}{AVGV-Huber} & \multirow{2}{*}{IDent} & \multirow{2}{*}{PIDent} \\
\multicolumn{2}{c}{}                             &                       &                             &                           &                           &                             &                           &                           &                             &                     &                       \\ \hline
\multicolumn{2}{c}{\multirow{8}{*}{Abalone}}     & \multirow{2}{*}{2}    &                             & 9.78                      & 5.05                      & 5.07                        & 9.78                      & 7.16                      & 7.23                        & \textbf{5.02}       & \textbf{5.02}         \\
\multicolumn{2}{c}{}                             &                       &                             & (0.57)                    & (0.30)                    & (0.29)                      & (0.57)                    & (0.55)                    & (0.56)                      & \textbf{(0.32)}     & \textbf{(0.30)}       \\
\multicolumn{2}{c}{}                             & \multirow{2}{*}{4}    &                             & 14.57                     & 5.44                      & 5.57                        & 14.57                     & 14.10                     & 14.05                       & 5.10                & \textbf{5.05}         \\
\multicolumn{2}{c}{}                             &                       & 5.02                        & (0.43)                    & (0.28)                    & (0.28)                      & (0.43)                    & (0.99)                    & (0.73)                      & (0.32)              & \textbf{(0.33)}       \\
\multicolumn{2}{c}{}                             & \multirow{2}{*}{8}    & (0.33)                       & 20.09                     & 7.91                      & 8.03                        & 20.09                     & 21.55                     & 20.43                       & 5.13                & \textbf{5.09}         \\
\multicolumn{2}{c}{}                             &                       &                             & (0.84)                    & (0.46)                    & (0.48)                      & (0.84)                    & (0.93)                    & (0.87)                      & (0.36)              & \textbf{(0.33)}       \\
\multicolumn{2}{c}{}                             & \multirow{2}{*}{16}   &                             & 25.12                     & 13.78                     & 13.94                       & 25.12                     & 27.21                     & 25.88                       & 5.25                & \textbf{5.16}         \\
\multicolumn{2}{c}{}                             &                       &                             & (0.67)                    & (0.96)                    & (0.70)                      & (0.67)                    & (0.87)                    & (0.68)                      & (0.34)              & \textbf{(0.35)}       \\ \hline
\multicolumn{2}{c}{\multirow{8}{*}{Airfoil}}     & \multirow{2}{*}{2}    &                             & 29.22                     & 23.78                     & 23.84                       & 29.22                     & 27.57                     & 27.74                       & \textbf{23.36}      & 23.38                 \\
\multicolumn{2}{c}{}                             &                       &                             & (2.47)                    & (1.98)                    & (2.05)                      & (2.47)                    & (2.30)                    & (2.43)                      & \textbf{(1.91)}     & (1.91)                \\
\multicolumn{2}{c}{}                             & \multirow{2}{*}{4}    &                             & 33.63                     & 25.10                     & 25.12                       & 33.63                     & 33.04                     & 33.28                       & 23.47               & \textbf{23.43}        \\
\multicolumn{2}{c}{}                             &                       & 23.22                       & (2.86)                    & (2.16)                    & (2.22)                      & (2.86)                    & (2.31)                    & (2.31)                      & (2.02)              & \textbf{(2.04)}       \\
\multicolumn{2}{c}{}                             & \multirow{2}{*}{8}    & (1.95)                      & 39.45                     & 29.13                     & 29.11                       & 39.45                     & 41.17                     & 40.62                       & 24.17               & \textbf{24.16}        \\
\multicolumn{2}{c}{}                             &                       &                             & (3.47)                    & (2.61)                    & (2.47)                      & (3.47)                    & (3.95)                    & (3.81)                      & (2.16)              & \textbf{(2.05)}       \\
\multicolumn{2}{c}{}                             & \multirow{2}{*}{16}   &                             & 44.36                     & 34.62                     & 34.75                       & 44.36                     & 45.91                     & 45.74                       & 24.74               & \textbf{24.59}        \\
\multicolumn{2}{c}{}                             &                       &                             & (3.99)                    & (3.41)                    & (3.48)                      & (3.99)                    & (4.41)                    & (4.11)                      & (2.33)              & \textbf{(2.26)}       \\ \hline
\multicolumn{2}{c}{\multirow{8}{*}{Auto-mpg}}    & \multirow{2}{*}{2}    &                             & 21.44                     & 11.67                     & 11.37                       & 21.38                     & 16.58                     & 16.13                       & 10.16               & \textbf{10.09}        \\
\multicolumn{2}{c}{}                             &                       &                             & (3.57)                    & (2.93)                    & (2.91)                      & (3.59)                    & (3.97)                    & (3.96)                      & (2.30)              & \textbf{(2.39)}       \\
\multicolumn{2}{c}{}                             & \multirow{2}{*}{4}    &                             & 31.57                     & 13.78                     & 13.64                       & 31.57                     & 31.01                     & 30.48                       & \textbf{11.04}      & 11.18                 \\
\multicolumn{2}{c}{}                             &                       & 10.05                       & (5.60)                    & (3.55)                    & (3.45)                      & (5.00)                    & (5.00)                    & (5.65)                      & \textbf{(3.36)}     & (3.33)                \\
\multicolumn{2}{c}{}                             & \multirow{2}{*}{8}    & (2.26)                      & 41.46                     & 18.45                     & 18.33                       & 41.46                     & 45.65                     & 43.81                       & 11.45               & \textbf{11.44}        \\
\multicolumn{2}{c}{}                             &                       &                             & (3.46)                    & (2.21)                    & (3.95)                      & (3.46)                    & (6.24)                    & (4.90)                      & (2.63)              & \textbf{(2.35)}       \\
\multicolumn{2}{c}{}                             & \multirow{2}{*}{16}   &                             & 51.06                     & 31.72                     & 32.06                       & 51.06                     & 56.10                     & 54.16                       & 12.73               & \textbf{12.59}        \\
\multicolumn{2}{c}{}                             &                       &                             & (6.86)                    & (6.24)                    & (6.09)                      & (6.86)                    & (5.94)                    & (6.74)                      & (3.16)              & \textbf{(2.88)}       \\ \hline
\multicolumn{2}{c}{\multirow{8}{*}{Housing}}     & \multirow{2}{*}{2}    &                             & 37.93                     & 26.30                     & \textbf{26.10}              & 37.93                     & 36.24                     & 35.28                       & 28.49               & 28.11                 \\
\multicolumn{2}{c}{}                             &                       &                             & (8.58)                    & (5.74)                    & \textbf{(5.79)}             & (8.58)                    & (8.61)                    & (8.31)                      & (6.88)              & (6.35)               \\
\multicolumn{2}{c}{}                             & \multirow{2}{*}{4}    &                             & 50.52                     & 31.17                     & 31.40                       & 50.52                     & 53.05                     & 51.51                       & 27.64               & \textbf{27.38}        \\
\multicolumn{2}{c}{}                             &                       & 27.30                       & (11.99)                   & (9.42)                    & 31.40                       & (11.99)                   & (13.48)                   & (12.87)                     & (7.15)              & \textbf{(6.53)}       \\
\multicolumn{2}{c}{}                             & \multirow{2}{*}{8}    & (5.99)                      & 60.85                     & 38.85                     & 38.67                       & 60.85                     & 63.91                     & 62.04                       & 32.15               & \textbf{31.78}        \\
\multicolumn{2}{c}{}                             &                       &                             & (9.83)                    & (8.77)                    & (8.91)                      & (9.83)                    & (8.79)                    & (9.23)                      & (10.11)             & \textbf{(10.37)}      \\
\multicolumn{2}{c}{}                             & \multirow{2}{*}{16}   &                             & 76.78                     & 52.62                     & 52.62                       & 76.78                     & 82.46                     & 80.25                       & 36.56               & \textbf{35.35}        \\
\multicolumn{2}{c}{}                             &                       &                             & (11.28)                   & (11.59)                   & (11.31)                     & (11.28)                   & (11.58)                   & (11.60)                     & (9.76)              & \textbf{(9.76)}       \\ \hline
\multicolumn{2}{c}{\multirow{8}{*}{Concrete}}    & \multirow{2}{*}{2}    &                             & 142.29                    & 115.93                    & 115.25                      & 142.29                    & 139.24                    & 138.41                      & 112.38              & \textbf{112.37}       \\
\multicolumn{2}{c}{}                             &                       &                             & (8.35)                    & (6.92)                    & (7.03)                      & (8.35)                    & (9.83)                    & (9.43)                      & (5.64)              & \textbf{(5.55)}       \\
\multicolumn{2}{c}{}                             & \multirow{2}{*}{4}    &                             & 175.40                    & 126.60                    & 126.32                      & 175.40                    & 182.35                    & 182.02                      & 112.18              & \textbf{112.11}       \\
\multicolumn{2}{c}{}                             &                       & 110.63                      & (11.53)                   & (6.53)                    & (6.70)                      & (11.53)                   & (14.66)                   & (14.35)                     & (7.22)              & \textbf{(7.33)}       \\
\multicolumn{2}{c}{}                             & \multirow{2}{*}{8}    & (5.36)                      & 209.66                    & 143.17                    & 143.40                      & 209.66                    & 222.84                    & 222.35                      & 115.23              & \textbf{114.77}       \\
\multicolumn{2}{c}{}                             &                       &                             & (16.07)                   & (8.05)                    & (7.91)                      & (16.07)                   & (16.14)                   & (15.76)                     & (6.78)              & \textbf{(7.39)}       \\
\multicolumn{2}{c}{}                             & \multirow{2}{*}{16}   &                             & 249.88                    & 196.43                    & 196.07                      & 249.88                    & 269.93                    & 268.21                      & \textbf{124.35}     & 133.28                \\
\multicolumn{2}{c}{}                             &                       &                             & (15.87)                   & (11.66)                   & (11.83)                     & (15.87)                   & (15.50)                   & (16.46)                     & \textbf{(6.75)}     & (27.76)               \\ \hline
\multicolumn{2}{c}{\multirow{8}{*}{Power-plant}} & \multirow{2}{*}{2}    &                             & 64.88                     & 23.91                     & 23.95                       & 64.88                     & 43.56                     & 43.85                       & 21.64               & \textbf{21.62}        \\
\multicolumn{2}{c}{}                             &                       &                             & (4.91)                    & (1.54)                    & (1.61)                      & (4.91)                    & (3.51)                    & (3.68)                      & (0.73)              & \textbf{(0.69)}       \\
\multicolumn{2}{c}{}                             & \multirow{2}{*}{4}    &                             & 104.63                    & 29.70                     & 29.66                       & 104.63                    & 99.26                     & 98.74                       & 21.60               & \textbf{21.53}        \\
\multicolumn{2}{c}{}                             &                       & 21.41                       & (4.34)                    & (1.71)                    & (1.73)                      & (4.34)                    & (7.38)                    & (6.36)                      & (0.95)              & \textbf{(0.95)}       \\
\multicolumn{2}{c}{}                             & \multirow{2}{*}{8}    & (0.64)                      & 154.75                    & 48.75                     & 48.99                       & 154.75                    & 169.72                    & 166.76                      & \textbf{21.55}      & 21.85                 \\
\multicolumn{2}{c}{}                             &                       &                             & (7.57)                    & (3.32)                    & (3.41)                      & (7.57)                    & (7.12)                    & (7.51)                      & \textbf{(0.70)}     & (0.77)                \\
\multicolumn{2}{c}{}                             & \multirow{2}{*}{16}   &                             & 202.64                    & 106.68                    & 106.91                      & 202.64                    & 225.97                    & 223.18                      & \textbf{22.02}      & 22.85                 \\
\multicolumn{2}{c}{}                             &                       &                             & (5.41)                    & (6.69)                    & (7.33)                      & (5.41)                    & (5.15)                    & (5.83)                      & \textbf{(0.79)}     & (2.09)                \\ \hline
\multicolumn{2}{c}{\multirow{8}{*}{Cpu-act}}     & \multirow{2}{*}{2}    &                             & 304.66                    & 139.01                    & 135.64                      & 304.66                    & 238.93                    & 238.35                      & 107.41              & \textbf{106.11}       \\
\multicolumn{2}{c}{}                             &                       &                             & (22.88)                   & 19.56                     & (17.08)                     & (22.88)                   & (13.04)                   & (12.48)                     & (10.86)             & \textbf{(9.55)}       \\
\multicolumn{2}{c}{}                             & \multirow{2}{*}{4}    &                             & 514.33                    & 143.62                    & 144.06                      & 514.33                    & 565.37                    & 561.82                      & 116.93              & \textbf{116.34}       \\
\multicolumn{2}{c}{}                             &                       & 98.24                       & (15.03)                   & (13.46)                   & (12.78)                     & (15.03)                   & (13.75)                   & (11.68)                     & (12.25)             & \textbf{(11.46)}      \\
\multicolumn{2}{c}{}                             & \multirow{2}{*}{8}    & (10.69)                     & 760.87                    & 371.09                    & 359.52                      & 774.19                    & 884.87                    & 858.18                      & 132.07              & \textbf{130.64}       \\
\multicolumn{2}{c}{}                             &                       &                             & (57.62)                   & (35.76)                   & (40.87)                     & (21.32)                   & (19.49)                   & 71.00                       & (15.29)             & \textbf{(15.35)}      \\
\multicolumn{2}{c}{}                             & \multirow{2}{*}{16}   &                             & 1000.14                   & 737.14                    & 762.46                      & 1000.14                   & 1125.31                   & 1131.31                     & \textbf{141.10}     & 143.12                \\
\multicolumn{2}{c}{}                             &                       &                             & (18.04)                   & (43.63)                   & (15.87)                     & (18.04)                   & (19.19)                   & (28.16)                     & \textbf{(12.60)}    & (15.92)              
\\
\bottomrule
\end{tabular}
}
\caption{Test performance (mean squared error with standard deviation) of each method on the seven benchmark datasets training with the Linear model. The best performance is highlighted in bold.}
\label{tab3}
\end{table*}
\paragraph{Candidate Label Set Generation.} Since this is the first work on PLR, there is no real-world PLR dataset where each instance is assigned a real-valued candidate label set. Hence we need to artificially generate candidate label sets by using the standard datasets in Table~\ref{tab1}. We assume that the generation of candidate label sets is instance-independent, which is a widely used data generation assumption in the weakly supervised learning field \cite{patrini2017making,ghosh2017robust,ishida2019complementary,feng2020provably}. We fix the size of the candidate label set and independently sample the false label multiple times to form the candidate label set. Formally speaking, let us denote by $\widetilde{y}$ a false label in the non-candidate label, then $\widetilde{y}$ is uniformly sampled from the empirically estimated span of label space in the training set (i.e., $\widetilde{y}\sim U(\min_{i\in [n]}y_i, \max_{i\in [n]}y_i)$). We adopt this uniform distribution because a larger candidate label set means more distractors, making the model more difficult to find the true label. For all the datasets, we denote by $|\bar{S}|$ the number of false labels in the candidate label set and set $|\bar{S}|$ to different values (including 2, 4, 8, and 16) for generating the candidate label set.

\paragraph{Compared Methods.}
In addition to the average method in Eq.~(\ref{avg}) that can serve as a baseline method, we also consider a variant of this method. Specifically, for each instance, we take the averaged value of all candidate labels as the true label, and minimize a conventional regression loss function to train the desired model. We name this variant of the average method AVGV and rename the average method AVGL. It is worth noting that the methods mentioned in this work can be equipped with arbitrary loss functions. Hence we substitute three regression losses into the AVGL method and the AVGV method, including the \emph{mean squared error} (MSE), the \emph{mean absolute error} (MAE), and the Huber loss. In this way, we can collect six baseline methods, including AVGL-MSE, AVGL-MAE, AVGL-Huber, AVGV-MSE, AVGV-MAE, and AVGV-Huber. We also compare with the supervised regression method that directly trains the model with MSE from fully labeled data (i.e., the true label is provided for each training instance). For the proposed identification method and progressive identification method, we rename them IDent and PIDent, and they are equipped with the commonly used MSE. For AVGL-Huber and AVGV-Huber, the threshold value of the Huber loss is selected from $\{1,5\}$. For our PIDent method, $\beta_1$ is fixed at $0.5$ and $\beta_2$ is selected from $\{10,100,500,1000,10000\}$. For all the methods, the learning rate is selected from $\{0.01,0.001\}$.

\begin{figure*}[!t]
\centering
  \subfigure[$|\bar{S}|=2$ on Concrete]{
      \includegraphics[width=1.5in]{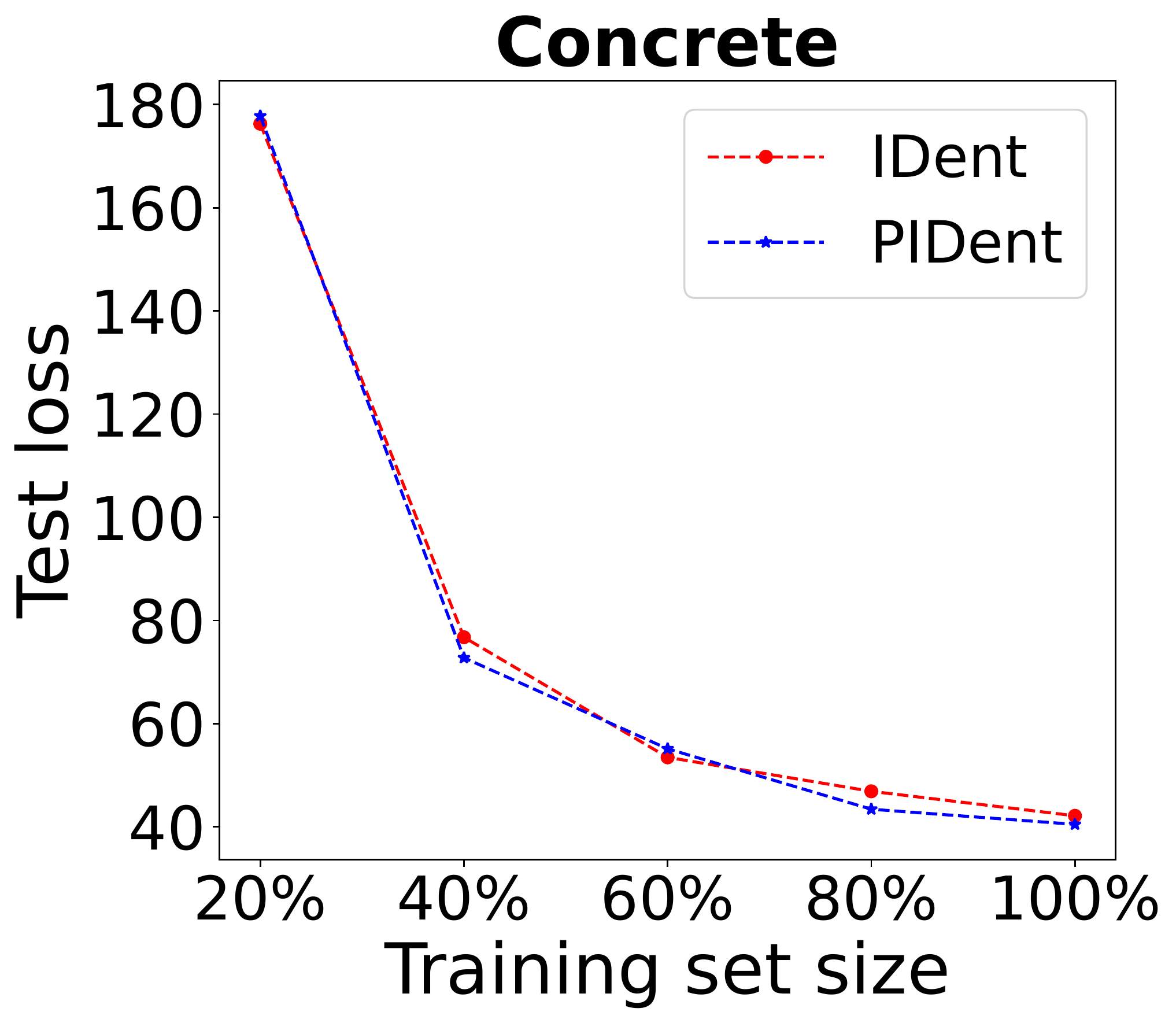}
      }
  \subfigure[$|\bar{S}|=4$ on Concrete]{
      \includegraphics[width=1.5in]{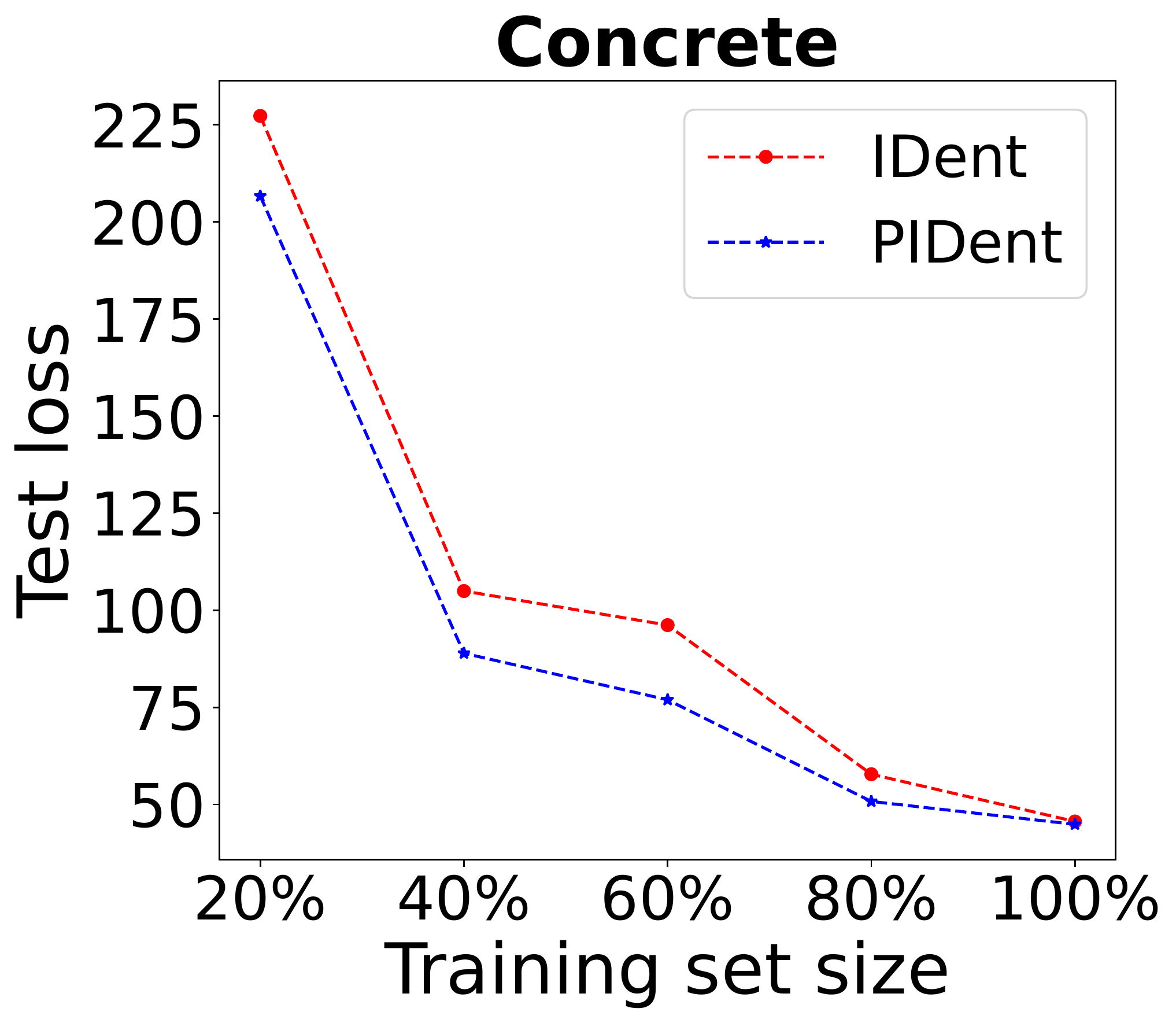}
      }
  \subfigure[$|\bar{S}|=2$ on Housing]{
      \includegraphics[width=1.5in]{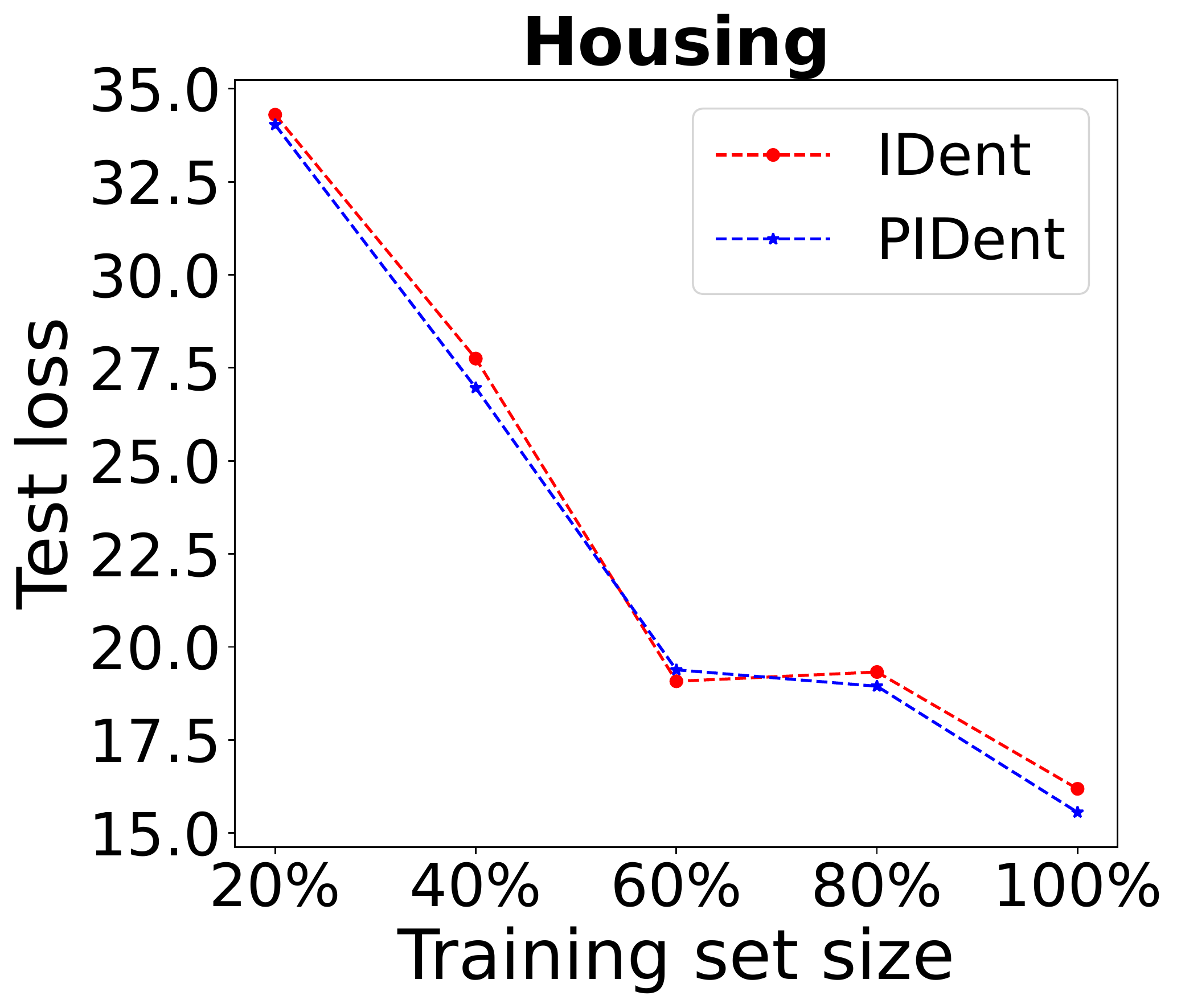}
      }
  \subfigure[$|\bar{S}|=4$ on Housing]{
      \includegraphics[width=1.5in]{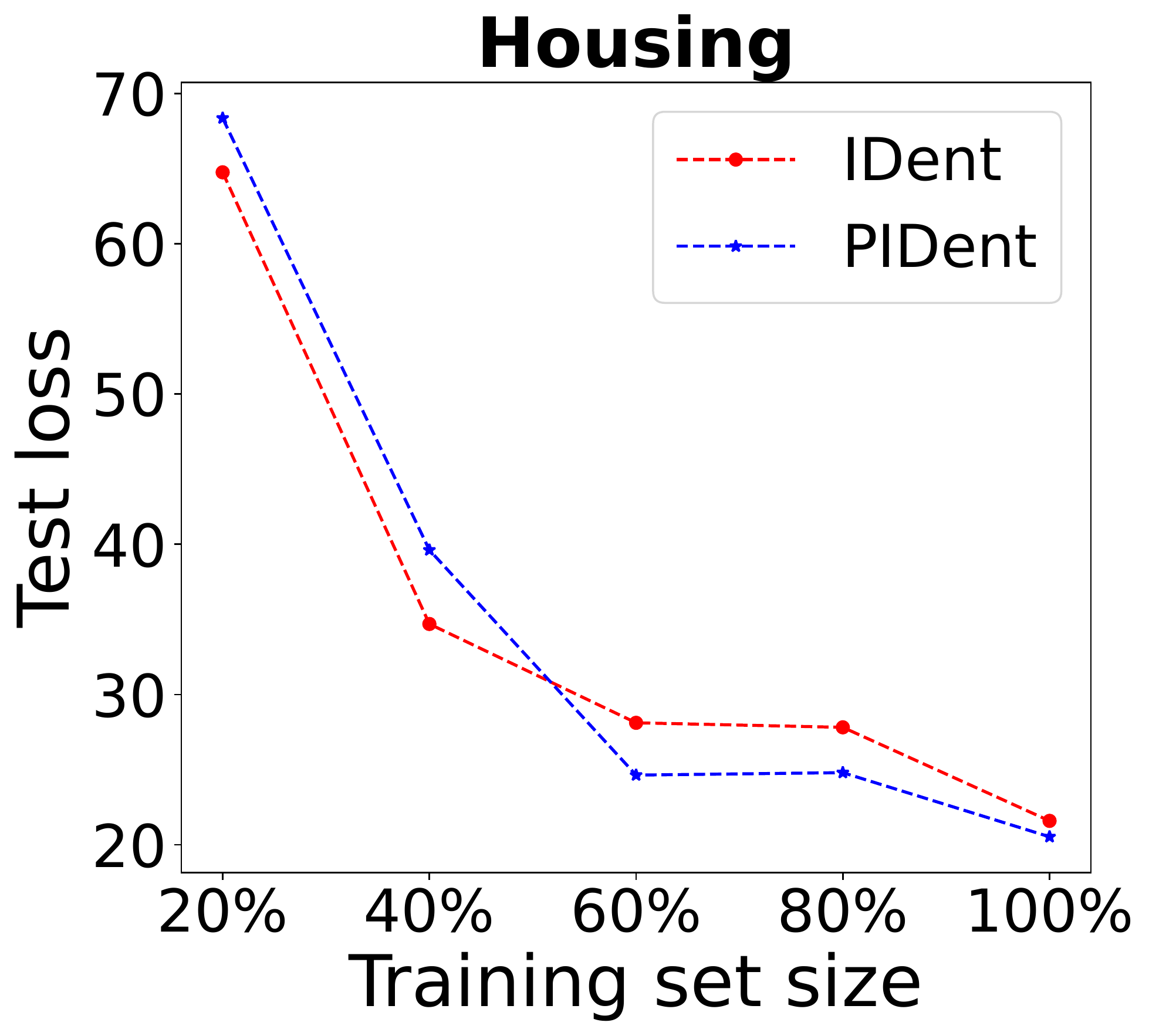}
      }
  \caption{The test performance on the Concrete and Housing datasets for the IDent method and the PIDent method when the number of training examples for partial-label regression increases.}
  \label{fig1} 
\end{figure*}
\subsection{Experimental Performance}
Table~\ref{tab2} and Table~\ref{tab3} show the mean squared error with standard deviation on the test set using the MLP model and the linear model, respectively. From the two tables, we have the following observations:
1) Our proposed methods IDent and PIDent outperform all the baseline methods, which demonstrates the effectiveness of the two methods. Besides, they could even be on a par with the supervised regression method in some cases, which verifies the ability of the two methods on identifying the true real-valued label.
2) As $|\bar{S}|$ increases, there exists a trend of degraded performance for all the PLR methods. This is because when the number of false labels in the candidate label set increases, more distracters will be included, and thus the PLR task will become more difficult. It is worth noting that with the increasing of $|\bar{S}|$, the test performance of all the baseline methods degrades dramatically while our proposed methods IDent and PIDent get worse only slightly.
3) By comparing the experimental results reported in Table \ref{tab2} and \ref{tab3}, we can observe that training with the MLP model is generally better than training with the linear model. This observation is in accordance with the common knowledge that the MLP model is more powerful than the linear model.
4) The PIDent method shows slightly better performance than the IDent method, which implies that the progressive identification strategy (i.e., soft weights) could be more promising than the direct identification strategy (i.e., hard pseudo-labeling). We may expect more significant improvements of the PIDent method over the IDent method with a better designed weighting function.
5) The AVGL methods are generally better than the AVGV methods, which implies that training with average loss is generally better than average value. Besides, MAE and Huber obviously outperform MSE, which shows that the robustness of MAE and Huber still holds in the PLR task.
\paragraph{Performance of Increasing PLR Training Data.} As we showed in Theorem \ref{thm2} and Theorem \ref{thm4}, the models learned by our proposed IDent method and the PIDent method could converge to the optimal model learned from fully labeled data when the number of PLR training examples approaches to infinity. Therefore, the performance of the two methods is expected to be improved if more PLR training examples are provided. To empirically validate such a theoretical finding, we further conduct experiments on the Concrete and Housing datasets by changing the fraction of PLR training examples (100\% means that we use all the PLR training examples in the training set). The experimental performance of the IDent method and the PIDent method when we increase the PLR training examples is provided in Figure \ref{fig1}. As shown in Figure~\ref{fig1}, the test loss of the two methods generally decreases when more PLR training examples are used for model training. This observation is clearly in accordance with our theoretical analyses in Theorem \ref{thm2} and Theorem \ref{thm4}, because the learned model would be closer to the optimal model as more PLR training examples are provided.
\section{Conclusion}
In this paper, we investigated a novel weakly supervised learning setting called partial-label regression, which is a variant of partial-label learning focusing on the regression task. 
To solve this problem, we first proposed a simple baseline method that takes the average loss incurred by candidate labels as the predictive loss. To overcome the drawback of this baseline method, we proposed an identification method that takes the least loss incurred by candidate labels as the predictive loss. We further propose a progressive identification method to differentiate candidate labels using progressively updated weights. We proved the model consistency of the latter two methods, which indicates that learned model can converge to the optimal model learned with fully labeled data. Finally, we conducted extensive experiments to demonstrate the effectiveness of our proposed methods. We expect that our first study with simple yet theoretically grounded methods for partial-label regression could inspire more research works on this new task.

We only used a uniform distribution to generate real-valued candidate label sets due to the page limit in the experiments. It would be interesting to further empirically investigate the performance of our proposed methods with other types of generation distributions. Besides, since our proposed methods can be compatible with any models, optimizers, and loss functions, they would be suited for dealing with large-scale regression datasets. Therefore, another promising direction is to apply our proposed methods to solve large-scale problems encountered in real-world application domains such as computer vision \cite{szeliski2010computer} and natural language processing \cite{chowdhary2020natural}. In addition, the performance of the PIDent method heavily relies on the designed weighting function. Hence how to properly design an effective weighting function for the PIDent method would be an interesting direction to further improve this method.

\section*{Acknowledgements}
This work is supported by the National Natural Science Foundation of China (Grant No. 62106028), Chongqing Overseas Chinese Entrepreneurship and Innovation Support Program, and CAAI-Huawei MindSpore Open Fund.

\bibliography{aaai23}
\appendix
\section{A\quad Proof of Theorem 1}
First, we prove that the optimal model $f^\star$ learned from fully labeled data (i.e., $f^\star = \argmin R(f)$) is also the optimal model for $R_{\mathrm{min}}(f)=\mathbb{E}_{p(\bm{x},S)}[\ell_{\mathrm{min}}(f(\bm{x}),S)]$ as follows. 

By substituting the $f^\star$ into $R_{\mathrm{min}}(f)$, we obtain:
\begin{align}
\nonumber
&R_{\mathrm{min}}(f^\star) \\
\nonumber
&=\mathbb{E}_{p(\bm{x}, S)}[\ell_{\mathrm{min}}(f^\star(\bm{x}),S)]\\
\nonumber
&=\int_{\mathcal{X}}\int_{\mathcal{S}}\ell_{\mathrm{min}}(f^\star(\bm{x}),S)p(S\mid\bm{x})p(\bm{x})\mathrm{d}S \mathrm{d}\bm{x}\\
\nonumber
&=\int_{\mathcal{X}}\int_{\mathcal{S}}\int_{\mathcal{Y}}\ell_{\mathrm{min}}(f^\star(\bm{x}),S)p(S,y\mid\bm{x})p(\bm{x})\mathrm{d}y \mathrm{d}S \mathrm{d}\bm{x}\\
\nonumber
&=\int_{\mathcal{X}}\int_{\mathcal{S}}\int_{\mathcal{Y}}\min\nolimits_{y^\prime\in S}\ell(f^\star(\bm{x}), y^\prime)p(S,y\mid\bm{x})p(\bm{x})\mathrm{d}y \mathrm{d}S \mathrm{d}\bm{x}\\
\nonumber
&=\int_{\mathcal{X}}\int_{\mathcal{Y}}\ell(f^\star(\bm{x}), y)\int_{\mathcal{S}}p(S\mid y,\bm{x})p(y\mid\bm{x})p(\bm{x})\mathrm{d}S \mathrm{d}y \mathrm{d}\bm{x}\\
\nonumber
&=\int_{\mathcal{X}}\int_{\mathcal{Y}}\ell(f^\star(\bm{x}), y)p(\bm{x},y)\mathrm{d}y \mathrm{d}\bm{x}\\
\nonumber
&=R(f^\star)=0,
\end{align}
which indicates that $f^\star$ is the optimal model for $R_{\mathrm{min}}(f)$. 

On the other hand, we prove that $f^\star$ is the sole optimal model for $R_{\mathrm{min}}(f)$ by contradiction. Specifically, we assume that there is at least one other model $g$ that makes $R_{\mathrm{min}}(g)=0$ and predicts a label $y_{g} \neq y$ for at least one instance $\bm{x}$. Therefore, for any $S$ containing $y$, we have
\begin{gather}
\nonumber
\min\nolimits_{y^\prime\in S}\ell(g(\bm{x}), y^\prime)=\ell(g(\bm{x}), y_{g})=0.
\end{gather}

The above equality implies that $y_{g}$ is always included in the candidate label set of $\bm{x}$ (co-occurring with the true label $y$), and in this case, the ambiguity degree is 1. This contradicts the basic PLR assumption that the ambiguity degree should be less than 1. Therefore, there is one, and only one minimizer of $R_{\mathrm{min}}$, which is the same as the minimizer $f^\star$ learned from fully labeled data. The proof is completed.\qed

\section{B\quad Proof of Theorem 2}
Let us introduce the following notations: 
\begin{align}
\nonumber
d &=\mathrm{Pdim}(\{\bm{x}\mapsto \ell(f(\bm{x}),y)\mid f\in\mathcal{F}\}),\\
\nonumber
d^\prime &= \mathrm{Pdim}(\{\bm{x}\mapsto \min_{y\in S}\ell(f(\bm{x}),y)\mid f\in\mathcal{F}\}),
\end{align}
where $\mathrm{Pdim}(\mathcal{F})$ denotes the pseudo-dimension of the functional space $\mathcal{F}$. It is worth noting that we may represent $d^\prime$ by $d$ with some derivations, while for simplicity and convenience, we directly formulate the expression of $d^\prime$.

From the assumptions in Theorem 2, using the discussion in Theorem 10.6 of \citet{mohri2012foundations}, with probability $1-\delta$ for all $f\in\mathcal{F}$,
\begin{align}
\nonumber
&\left|\mathbb{E}_{p(\bm{x},S)}[\ell_{\mathrm{min}}(\bm{x},S)] - \sum_{i=1}^n\ell_{\mathrm{min}}(\bm{x}_i,S_i)\right| \\
\nonumber
&\qquad\qquad\qquad\qquad\qquad\leq M\sqrt{\frac{2d^\prime\log\frac{ne}{d^\prime}}{n}}+M\sqrt{\frac{\log\frac{2}{\delta}}{2n}}.
\end{align}
Hence we can know that $\left|R_{\mathrm{min}}(f)-\widehat{R}_{\mathrm{min}}(f)\right|$ converges in the order of $\mathcal{O}(1/\sqrt{n})$ for all $f\in\mathcal{F}$.

Then, we have the following derivations:
\begin{align}
\nonumber
&R_{\mathrm{min}}(\widehat{f}_{\mathrm{min}})-R_{\mathrm{min}}(f^\star)\\
\nonumber
&\leq R_{\mathrm{min}}(\widehat{f}_{\mathrm{min}}) - \widehat{R}_{\mathrm{min}}(\widehat{f}_{\mathrm{min}}) + \widehat{R}_{\mathrm{min}}(\widehat{f}_{\mathrm{min}}) - \widehat{R}_{\mathrm{min}}(f^\star) \\
\nonumber
&\qquad\qquad\qquad\qquad\qquad\qquad
+ \widehat{R}_{\mathrm{min}}(f^\star) - R_{\mathrm{min}}(f^\star)\\
\nonumber
&\leq R_{\mathrm{min}}(\widehat{f}_{\mathrm{min}}) - \widehat{R}_{\mathrm{min}}(\widehat{f}_{\mathrm{min}}) + \widehat{R}_{\mathrm{min}}(f^\star) - R_{\mathrm{min}}(f^\star)\\
\nonumber
&\leq \left|R_{\mathrm{min}}(\widehat{f}_{\mathrm{min}}) - \widehat{R}_{\mathrm{min}}(\widehat{f}_{\mathrm{min}})\right|\\
\nonumber
&\qquad\qquad\qquad\qquad\qquad\qquad+ \left|\widehat{R}_{\mathrm{min}}(f^\star) - R_{\mathrm{min}}(f^\star)\right|,
\end{align}
where the first inequality holds because $\widehat{R}_{\mathrm{min}}(\widehat{f}_{\mathrm{min}}) - \widehat{R}_{\mathrm{min}}(f^\star)\leq 0$ since $\widehat{f}_{\mathrm{min}}=\argmin_{f\in\mathcal{F}}\widehat{R}_{\mathrm{min}}(f)$. As both the last two terms in the last line converges in the order of $\mathcal{O}(1/\sqrt{n})$, we can know that $R_{\mathrm{min}}(\widehat{f}_{\mathrm{min}})-R_{\mathrm{min}}(f^\star)$ converges in the order of $\mathcal{O}(1/\sqrt{n})$.
The proof is completed.\qed
\section{C\quad Proof of Theorem 3}
First, we prove that the optimal model $f^\star$ learned from fully labeled data (i.e., $f^\star = \argmin R(f)$) is also the optimal model for $R_{\mathrm{wet}}(f)=\mathbb{E}_{p(\bm{x},S)}[\ell_{\mathrm{wet}}(f(\bm{x}),S)]$ as follows. 

By substituting the $f^\star$ into $R_{\mathrm{wet}}(f)$, we obtain:
\begin{align}
\nonumber
&R_{\mathrm{wet}}(f^\star) \\
\nonumber
&=\mathbb{E}_{p(\bm{x}, S)}[\ell_{\mathrm{wet}}(f^\star(\bm{x}),S)]\\
\nonumber
&=\int_{\mathcal{X}}\int_{\mathcal{S}}\ell_{\mathrm{wet}}(f^\star(\bm{x}),S)p(S\mid\bm{x})p(\bm{x})\mathrm{d}S \mathrm{d}\bm{x}\\
\nonumber
&=\int_{\mathcal{X}}\int_{\mathcal{S}}\int_{\mathcal{Y}}\ell_{\mathrm{wet}}(f^\star(\bm{x}),S)p(S,y\mid\bm{x})p(\bm{x})\mathrm{d}y \mathrm{d}S \mathrm{d}\bm{x}\\
\nonumber
&=\int_{\mathcal{X}}\int_{\mathcal{S}}\int_{\mathcal{Y}}\sum\nolimits_{y^\prime\in S}w(\bm{x},y^\prime)\ell(f^\star(\bm{x}), y^\prime)\\
\nonumber
&\qquad\qquad\qquad\qquad\qquad\qquad\quad\cdot p(S,y\mid\bm{x})p(\bm{x})\mathrm{d}y \mathrm{d}S \mathrm{d}\bm{x}\\
\nonumber
\nonumber
&=\int_{\mathcal{X}}\int_{\mathcal{Y}}\ell(f^\star(\bm{x}), y)\int_{\mathcal{S}}p(S\mid y,\bm{x})p(y\mid\bm{x})p(\bm{x})\mathrm{d}S \mathrm{d}y \mathrm{d}\bm{x}\\
\nonumber
&=\int_{\mathcal{X}}\int_{\mathcal{Y}}\ell(f^\star(\bm{x}), y)p(\bm{x},y)\mathrm{d}y \mathrm{d}\bm{x}\\
\nonumber
&=R(f^\star)=0,
\end{align}
where we used the equality $\sum\nolimits_{y^\prime\in S}w(\bm{x},y^\prime)\ell(f^\star(\bm{x}), y^\prime)=\ell(f^\star(\bm{x}),y)$. This is because when the true label $y\in S$ can make $\ell(f^\star(\bm{x}),y)=0$, and thus the weighting function would be $w(\bm{x},y)=1$ and $w(\bm{x},y^\prime)=0$ for $y^\prime\neq y$ since we try to minimize $\ell_{\mathrm{wet}}$.

On the other hand, we prove that $f^\star$ is the sole optimal model for $R_{\mathrm{wet}}(f)$ by contradiction. Specifically, we assume that there is at least one other model $g$ that makes $R_{\mathrm{wet}}(g)=0$ and predicts a label $y_{g} \neq y$ for at least one instance $\bm{x}$. Therefore, for any $S$ containing $y$ we have
\begin{gather}
\nonumber
\sum\nolimits_{y^\prime\in S}w(\bm{x},y^\prime)\ell(g(\bm{x}), y^\prime)=\ell(g(\bm{x}), y_{g})=0.
\end{gather}
The above equality implies that $y_{g}$ is always included in the candidate label set of $\bm{x}$ (co-occurring with the true label $y$), and in this case, the ambiguity degree is 1. This contradicts the basic PLR assumption that the ambiguity degree should be less than 1. Therefore, there is one, and only one minimizer of $R_{\mathrm{wet}}$, which is the same as the minimizer $f^\star$ learned from fully labeled data. The proof is completed.\qed

\section{D\quad Proof of Theorem 4}
Let us introduce the following notations: 
\begin{align}
\nonumber
d &=\mathrm{Pdim}(\{\bm{x}\mapsto \ell(f(\bm{x}),y)\mid f\in\mathcal{F}\}),\\
\nonumber
\tilde{d} &= \mathrm{Pdim}(\{\bm{x}\mapsto \sum_{y\in S}w(\bm{x},y)\ell(f(\bm{x}),y)\mid f\in\mathcal{F}\}),
\end{align}
where $\mathrm{Pdim}(\mathcal{F})$ denotes the pseudo-dimension of the functional space $\mathcal{F}$ and $w(\bm{x},y)$ satisfy the basic conditions described in the main text. It is worth noting that we may represent $\tilde{d}$ by $d$ with some derivations, while for simplicity and convenience, we directly formulate the expression of $\tilde{d}$.

From the assumptions in Theorem 2, using the discussion in Theorem 10.6 of \citet{mohri2012foundations}, with probability $1-\delta$ for all $f\in\mathcal{F}$,
\begin{align}
\nonumber
&\left|\mathbb{E}_{p(\bm{x},S)}[\ell_{\mathrm{wet}}(\bm{x},S)] - \sum_{i=1}^n\ell_{\mathrm{wet}}(\bm{x}_i,S_i)\right| \\
\nonumber
&\qquad\qquad\qquad\qquad\qquad\leq M\sqrt{\frac{2\tilde{d}\log\frac{ne}{\tilde{d}}}{n}}+M\sqrt{\frac{\log\frac{2}{\delta}}{2n}}.
\end{align}
Hence we can know that $\left|R_{\mathrm{wet}}(f)-\widehat{R}_{\mathrm{wet}}(f)\right|$ converges in the order of $\mathcal{O}(1/\sqrt{n})$ for all $f\in\mathcal{F}$.

Then, we have the following derivations:
\begin{align}
\nonumber
&R_{\mathrm{wet}}(\widehat{f}_{\mathrm{wet}})-R_{\mathrm{wet}}(f^\star)\\
\nonumber
&\leq R_{\mathrm{wet}}(\widehat{f}_{\mathrm{wet}}) - \widehat{R}_{\mathrm{wet}}(\widehat{f}_{\mathrm{wet}}) + \widehat{R}_{\mathrm{wet}}(\widehat{f}_{\mathrm{wet}}) - \widehat{R}_{\mathrm{wet}}(f^\star) \\
\nonumber
&\qquad\qquad\qquad\qquad\qquad\qquad
+ \widehat{R}_{\mathrm{wet}}(f^\star) - R_{\mathrm{wet}}(f^\star)\\
\nonumber
&\leq R_{\mathrm{wet}}(\widehat{f}_{\mathrm{wet}}) - \widehat{R}_{\mathrm{wet}}(\widehat{f}_{\mathrm{wet}}) + \widehat{R}_{\mathrm{wet}}(f^\star) - R_{\mathrm{wet}}(f^\star)\\
\nonumber
&\leq \left|R_{\mathrm{wet}}(\widehat{f}_{\mathrm{wet}}) - \widehat{R}_{\mathrm{wet}}(\widehat{f}_{\mathrm{wet}})\right|\\
\nonumber
&\qquad\qquad\qquad\qquad\qquad\qquad+ \left|\widehat{R}_{\mathrm{wet}}(f^\star) - R_{\mathrm{wet}}(f^\star)\right|,
\end{align}
where the first inequality holds because $\widehat{R}_{\mathrm{wet}}(\widehat{f}_{\mathrm{wet}}) - \widehat{R}_{\mathrm{wet}}(f^\star)\leq 0$ since $\widehat{f}_{\mathrm{wet}}=\argmin_{f\in\mathcal{F}}\widehat{R}_{\mathrm{wet}}(f)$. As both the last two terms in the last line converges in the order of $\mathcal{O}(1/\sqrt{n})$, we can know that $R_{\mathrm{wet}}(\widehat{f}_{\mathrm{wet}})-R_{\mathrm{wet}}(f^\star)$ converges in the order of $\mathcal{O}(1/\sqrt{n})$.
The proof is completed.\qed
\end{document}